\def\year{2016}\relax
\documentclass[letterpaper]{article}
\usepackage{aaai16}
\usepackage{times}
\usepackage{helvet}
\usepackage{courier}

\usepackage{amsmath,amssymb,amstext,amsthm}

\usepackage{booktabs}
\usepackage{graphics}
\usepackage{algorithmic,algorithm}
\usepackage{graphicx}
\usepackage{scalerel}

\nocopyright

\newcommand{\true}{\textrm{true}}
\newcommand{\avg}{\textrm{avg}}
\newcommand{\worst}{\textrm{worst}}

\newcommand{\MAP}{\textrm{MAP}}
\newcommand{\defeq}{\mathrel{\raisebox{-0.3ex}{\ensuremath{\stackrel{_{\rm def}}{=}}}}}
\newcommand{\itemset}{\mathcal{X}}

\newcommand{\labelset}{\mathcal{Y}}

\newtheorem{theorem}{Theorem}
\newtheorem{corollary}{Corollary}

\begin{document}
%

\title{Robustness of Bayesian Pool-based Active Learning Against Prior Misspecification}
\author{Nguyen Viet Cuong$^1$ \qquad\qquad Nan Ye$^2$ \qquad\qquad Wee Sun Lee$^3$\\ \\
$^1$Department of Mechanical Engineering, National University of Singapore, Singapore, \it{nvcuong@nus.edu.sg}\\
$^2$Mathematical Sciences School \& ACEMS, Queensland University of Technology, Australia, \it{n.ye@qut.edu.au}\\
$^3$Department of Computer Science, National University of Singapore, Singapore, \it{leews@comp.nus.edu.sg}
}

\maketitle

\begin{abstract}
\begin{quote}
  We study the robustness of active learning (AL) algorithms against
  prior misspecification: whether an algorithm achieves similar
  performance using a perturbed prior as compared to using the true
  prior.  In both the average and worst cases of the maximum coverage
  setting, we prove that all $\alpha$-approximate algorithms are robust
  (i.e., near $\alpha$-approximate) if the utility is Lipschitz
  continuous in the prior.  We further show that robustness may not be
  achieved if the utility is non-Lipschitz.  This suggests we should
  use a Lipschitz utility for AL if robustness is required. 
  For the minimum cost setting, we can also obtain a robustness result for approximate AL algorithms.
  Our results imply that many commonly used AL algorithms are robust against
  perturbed priors.  We then propose the use of a mixture prior
  to alleviate the problem of prior misspecification.  We analyze the
  robustness of the uniform mixture prior and show experimentally that it
  performs reasonably well in practice.
\end{quote}
\end{abstract}

\section{Introduction}
In pool-based active learning (AL), training examples are sequentially
selected and labeled from a pool of unlabeled data, with the aim of
obtaining a good classifier using as few labeled examples as possible
\cite{mccallum1998employing}.  To achieve computational efficiency,
most commonly used methods greedily select one example at a time based
on some criterion.

In this paper, we consider Bayesian pool-based AL that assumes data
labels are generated from a prior distribution.  In theory, the true
prior is generally assumed to be known
\cite{golovin2011adaptive,cuong2013active,cuong2014near}.  In
practice, it is often unknown and misspecified; that is, the prior
used is different from the true one.  This work is the first one
investigating the \emph{robustness} of AL algorithms against prior
misspecification -- that is, whether an algorithm achieves similar
performance using a perturbed prior as compared to using the true prior.

We focus on the analysis of approximate algorithms for two commonly
studied problems: the \emph{maximum coverage problem} which aims to
maximize the expected or worst-case utility of the chosen examples given a fixed
budget of queries, and the \emph{minimum cost problem} which aims to
minimize the expected number of queries needed to identify the true
labeling of all examples. We focus on approximate algorithms
because previous works have shown that, in general, it is
computationally intractable to find the optimal strategy for choosing the
examples, while some commonly used AL algorithms can achieve good
approximation ratios compared to the optimal strategies
\cite{golovin2011adaptive,chen13near,cuong2013active,cuong2014near}.
For example, with the version space reduction utility, the maximum
Gibbs error algorithm achieves a $(1-1/e)$-approximation of the
optimal expected utility \cite{cuong2013active}, while the least
confidence algorithm achieves the same approximation of the optimal
worst-case utility \cite{cuong2014near}.

Our work shows that many commonly used AL algorithms are robust.  In the
maximum coverage setting, our main result is that if the utility
function is Lipschitz continuous in the prior, all $\alpha$-approximate
algorithms are robust, i.e., they are near $\alpha$-approximate when using
a perturbed prior.  More precisely, their performance guarantee on the
expected or worst-case utility degrades by at most a constant factor
of the $\ell_1$ distance between the perturbed prior and the true
prior.  It follows from this result that the maximum Gibbs error and
the least confidence algorithms are near $(1-1/e)$-approximate.  Our
result also implies the robustness of the batch and generalized
versions of the maximum Gibbs error algorithm.  If the utility is
non-Lipschitz, we show that even an optimal algorithm for the
perturbed prior may not be robust.  This suggests we should use a
Lipschitz continuous utility for AL in order to achieve robustness. 
Similarly, we prove a robustness
result for the minimum cost setting that implies the
robustness of the generalized binary search AL algorithm
\cite{dasgupta2004analysis,nowak2008generalized}.

We also address the difficulty of choosing a good prior in practice.
Practically, it is often easier to come up with a set of distributions and combine them using a mixture. 
Our theoretical results imply robustness when the mixture prior is close to the true prior. 
In the mixture setting, another interesting question is robustness when the true prior is one of the components of the mixture. 
In this case, the mixture prior may not necessarily be close to the true prior in terms of $\ell_1$ distance. 
We prove that for the uniform mixture prior, approximate AL algorithms are still robust in the sense that they are competitive with the optimum performance of the mixture, 
which is the performance we expect when modeling.
Our experiments show that the uniform mixture performs well in practice.

{\bf Related Works:}
Greedy algorithms for pool-based AL
usually optimize some measure of uncertainty of the selected examples
\cite{settles2010active,cuong2013active}.
In the Bayesian setting, they can be viewed as
greedily optimizing some corresponding average-case or worst-case objective.
For instance, the maximum entropy algorithm \cite{settles2008analysis}, 
which maximizes the Shannon entropy of the selected examples,
attempts to greedily maximize the policy entropy in the average case \cite{cuong2013active}.
The maximum Gibbs error algorithm, which maximizes the Gibbs error
of the selected examples, 
attempts to greedily maximize the version space reduction in the average case \cite{cuong2013active};
and the least confidence algorithm, which minimizes the probability of the most likely label of the selected examples, 
attempts to maximize the version space reduction in the worst case \cite{cuong2014near}.

Analyses of these algorithms typically investigate their
near-optimality guarantees in the average or worst case.  The maximum
entropy algorithm was shown to have no constant factor approximation guarantee
in the average case \cite{cuong2014near}.  In contrast, the maximum
Gibbs error algorithm has a $(1-1/e)$-factor approximation guarantee
for the average version space reduction objective
\cite{cuong2013active}.  This algorithm is a probabilistic version of
the generalized binary search algorithm
\cite{dasgupta2004analysis,golovin2011adaptive}.  It can also be
applied to the batch mode setting \cite{hoi2006batch}, and was shown
to provide a $(1 - e^{-(e-1)/e})$-factor approximation to the optimal
batch AL algorithm \cite{cuong2013active}.  In the noiseless case,
this batch maximum Gibbs error algorithm is equivalent to the
BatchGreedy algorithm \cite{chen13near}.

\citeauthor{cuong2014near} \shortcite{cuong2014near} showed 
the least confidence algorithm \cite{culotta2005reducing} has a
 $(1-1/e)$-factor approximation guarantee with respect to 
the worst-case version space reduction objective. 
A similar result in the worst case was also shown
for the generalized maximum Gibbs error algorithm with an arbitrary loss \cite{cuong2014near}.
These results are due to the pointwise submodularity of version space reduction. 
AL that exploits submodularity was also studied in \cite{guillory2010interactive,wei2015-submodular-data-active}.

\section{Preliminaries}
Let $\itemset$ be a finite set (a pool) of examples and $\labelset$ be a finite set of labels.
Consider the hypothesis space $\mathcal{H} \defeq \labelset ^ \itemset$
consisting of all functions from $\itemset$ to $\labelset$. 
Each hypothesis $h \in \mathcal{H}$ is a \emph{labeling} of $\itemset$.
In the Bayesian setting, we assume an unknown true labeling
$h_{\true}$ drawn from a prior $p_0[h]$ on $\mathcal{H}$.
After observing a labeled set $\mathcal{D}$, we obtain the posterior
$p_{\mathcal{D}}[h] \defeq p_0[h \mid \mathcal{D}]$ using Bayes' rule.

The true labeling $h_{\true}$ may be generated by a complex process
rather than directly drawn from a prior $p_0$.
For instance, if probabilistic models (e.g., naive Bayes) are used 
to generate labels for the examples
and a prior is imposed on these models instead of the labelings, we can convert this prior
into an equivalent prior on the labelings and work with this induced prior.
The construction of the induced prior involves computing the probability of labelings 
with respect to the original prior \cite{cuong2013active}.
In practice, we do not need to compute or maintain the induced prior explicitly as this process is very expensive.
Instead, we compute or approximate the AL criteria directly from the original prior on the probabilistic models \cite{cuong2013active,cuong2014near}.

For any distribution $p[h]$, any example sequence $S \subseteq \itemset$, 
and any label sequence $\mathbf{y}$ with the same length, 
$p[\mathbf{y};S]$ denotes the probability that $\mathbf{y}$ is the label sequence of $S$.
Formally, ${ p[\mathbf{y};S] \defeq \sum_h p[h] \, \mathbb{P}[h(S) = \mathbf{y} \mid h] }$,
where $h(S) = (h(x_1), \ldots, h(x_i))$ if $S = (x_1, \ldots, x_i)$.
In the above, $\mathbb{P}[h(S) = \mathbf{y} \mid h]$ is the probability that
$S$ has label sequence $\mathbf{y}$ given the hypothesis $h$.
If $h$ is deterministic as in our setting,
$\mathbb{P}[h(S) = \mathbf{y} \mid h] = \mathbf{1}(h(S)=\mathbf{y})$, where $\mathbf{1}(\cdot)$ is the indicator function.
Note that $p[\, \cdot \,;S]$ is a probability distribution on the label sequences of $S$.
We also write $p[y;x]$ to denote $p[\{ y \}; \{ x \} ]$ for $x \in \itemset$ and $y \in \labelset$.

Given a prior, a pool-based AL algorithm is equivalent 
to a policy for choosing training examples from $\itemset$.
A policy is a mapping from a partial labeling
(labeling of a subset of $\itemset$) to the next unlabeled example to query.
It can be represented by a policy tree whose nodes correspond to 
unlabeled examples to query, and edges from a node correspond to its labels.
When an unlabeled example is queried, 
we receive its true label according to $h_{\true}$.
We will focus on adaptive policies, which use the observed labels of previously chosen examples to query the next unlabeled example.

\section{Robustness: Maximum Coverage Problem}
\label{sec:robust}

We now consider the robustness of AL algorithms for the \emph{maximum coverage} problem:
find an adaptive policy maximizing the expected or worst-case utility given a budget of $k$ queries \cite{cuong2013active,cuong2014near}.
The \emph{utility} is a non-negative function 
$f(S,h) : 2^{\itemset} \times \mathcal{H} \rightarrow \mathbb{R}_{\geq 0}$. 
Intuitively, a utility measures the value of querying examples $S$
when the true labeling is $h$.
Utilities for AL usually depend on the prior, 
so we shall use the notation $f_p(S,h)$ to denote that the utility $f_p$ depends on a distribution $p$ over $\mathcal{H}$.

$f_p$ is said to be \emph{Lipschitz continuous (in the
prior)} with a Lipschitz constant $L$ if for any $S$, $h$, and two
priors $p$, $p'$,
\begin{equation}
|f_{p}(S, h) - f_{p'}(S, h)| \le L \|p - p'\|, \label{eq:Lipschitz}
\end{equation}
where $\|p - p'\| \hspace{-0.2em}\defeq\hspace{-0.2em} \sum_{h} | p[h] - p'[h] |$ is the $\ell_1$
distance between $p$ and $p'$.
Lipschitz continuity implies boundedness.\footnote{Choose an arbitrary $p'$, then for any $p, S, h$, we have
$f_{p}(S, h) 
\le f_{p'}(S, h) + L \|p - p'\| 
\le f_{p'}(S, h) + 2L
\le \max_{S, h} f_{p'}(S, h) + 2L$.
Similarly, a lower bound exists.
}

An AL algorithm is a mapping from a utility and a prior to a policy.
Let $x^\pi_h$ denote the set of examples selected by a policy $\pi$ when
the true labeling is $h$.
We now analyze the robustness of AL algorithms for both the
average case and the worst case.

\subsection{The Average Case}
\label{sec:average}

In this case, our objective is to find a policy with maximum expected utility.
If $p_0$ is the true prior, the expected utility of a policy $\pi$ is
$f^{\avg}_{p_0}(\pi) \defeq \mathbb{E}_{h \sim p_0} \left[ f_{p_0}(x^\pi_h, h) \right]$.

We consider the case where we have already chosen a utility, 
but still need to choose the prior.
In practice, the choice is often subjective and may not be the true prior.
A natural question is: if we choose a \emph{perturbed} prior $p_1$ 
(i.e., a prior not very different from the true prior $p_0$ in terms of $\ell_1$ distance), 
can an AL algorithm achieve performance
competitive to that obtained using the true prior?

Our first robustness result is for \emph{$\alpha$-approximate} algorithms
that return an $\alpha$-approximate policy of the optimal one.
Formally, an \emph{average-case $\alpha$-approximate} ($0 < \alpha \le 1$) algorithm $A$ outputs, for any prior $p$, a policy $A(p)$ satisfying
\[
f^{\avg}_{p}(A(p)) \ge \alpha \max_{\pi} f^{\avg}_{p}(\pi).
\]
When $\alpha = 1$, we call $A$ an exact algorithm.
For notational convenience, we drop the dependency of $A$ on the
utility as we assumed a fixed utility here.
We consider approximate algorithms because 
practical algorithms are generally approximate due to
computational intractability of the problem.
We have the following robustness result.
The proof of this theorem is given in Appendix \ref{apd-proof:theorem:average-active}.

\begin{theorem}
\label{theorem:average-active}
Assume the utility $f_{p}$ is Lipschitz continuous with a Lipschitz constant
$L$.
Let $M$ be an upper bound of $f_{p}$.
If $A$ is an average-case $\alpha$-approximate algorithm,
then for any true prior $p_0$ and any perturbed prior $p_1$,
\begin{equation*}
f^{\avg}_{p_0}(A(p_1)) \geq \alpha \max_{\pi} f^{\avg}_{p_0}(\pi) - (\alpha + 1) (L + M) \| p_1 - p_0 \|. 
\end{equation*}
Thus, $A$ is robust in the sense that it returns a near $\alpha$-approximate policy when
using a perturbed prior.
\end{theorem}

$f^{\avg}_{p_0}(A(p_1))$ is the expected utility of the policy returned by $A$ using $p_1$ as prior.
The expected utility is always computed with respect to the true prior $p_0$.
Theorem \ref{theorem:average-active} shows that when we use a perturbed prior $p_1$,
the expected utility achieved by an average-case $\alpha$-approximate algorithm degrades by at most a constant
factor of the $\ell_1$ distance between the perturbed prior and the true prior.

{\bf Application to Maximum Gibbs Error:}
Theorem \ref{theorem:average-active} implies the robustness of
the maximum Gibbs error algorithm \cite{cuong2013active}.
This algorithm greedily selects the next example $x^*$ satisfying
${ x^* = \arg \max_x \mathbb{E}_{y \sim p_{\mathcal{D}}[\cdot;x]} [ 1 - p_{\mathcal{D}} [y;x] ] }$,
where $p_{\mathcal{D}}$ is the current posterior and $p_{\mathcal{D}} [y;x]$ is
the probability (w.r.t. $p_{\mathcal{D}}$) that $x$ has label $y$.
In the binary-class and noiseless setting, it is equivalent to the 
generalized binary search algorithm \cite{dasgupta2004analysis,nowak2008generalized,golovin2011adaptive}.
Consider the version space reduction utility
$f_p(S, h) \defeq 1 - p[h(S);S]$, where $p[h(S);S]$ is the probability (w.r.t. $p$) that
$S$ has the labels $h(S)$.
We have the following corollary about the algorithm.
The proof of this corollary is given in Appendix \ref{apd-proof:corollary:maxGEC}.

\begin{corollary}
\label{corollary:maxGEC}
If $A$ is the maximum Gibbs error algorithm, then for any true prior $p_0$ and
any perturbed prior $p_1$,
\begin{equation*}
f^{\avg}_{p_0}(A(p_1)) \ge \Big(1 - \frac{1}{e} \Big) \max_{\pi} f^{\avg}_{p_0}(\pi) - \Big( 4 - \frac{2}{e} \Big) \| p_1 - p_0 \|.
\end{equation*}
\end{corollary}

{\bf Application to Batch Maximum Gibbs Error:}
We can also obtain the robustness result for the batch version of 
the maximum Gibbs error algorithm.
In the batch setting, the AL algorithm queries a batch of
examples in each iteration instead of only one example \cite{hoi2006batch}.
The batch maximum Gibbs error algorithm is
described in Algorithm 1 of \cite{cuong2013active},
and by Theorem 5 of the same work, it is a $(1 - e^{-(e-1)/e})$-approximate algorithm for the version space reduction utility above.
If we restrict the policies to only those in the batch setting, then from
Theorem \ref{theorem:average-active}, we have the following corollary.
Note that the range of the $\max$ operation in the corollary is restricted to only batch policies.

\begin{corollary}
\label{corollary:batch-maxGEC}
If $A$ is the batch maximum Gibbs error algorithm, for any true prior $p_0$ and
any perturbed prior $p_1$, 
\begin{align*}
f^{\avg}_{p_0}(A(p_1)) \ge& \left( 1 - e^{-(e-1)/e} \right) \max_{\pi} f^{\avg}_{p_0}(\pi) \\
 & - \left( 4 - 2 e^{-(e-1)/e} \right) \| p_1 - p_0 \|.
\end{align*}
\end{corollary}

\subsection{The Worst Case}

In this case, our objective is to find a policy with maximum worst-case utility.
If $p_0$ is the true prior, the worst-case utility of a policy $\pi$ is
$f^{\worst}_{p_0}(\pi) \defeq \min_h \left[ f_{p_0}(x^\pi_h, h) \right]$.

An algorithm $A$ is a \emph{worst-case $\alpha$-approximate} algorithm 
(${ 0 < \alpha \le 1 }$) if for any prior $p$, we have
${ f^{\worst}_{p}(A(p)) \ge \alpha \max_{\pi} f^{\worst}_{p}(\pi) }$.
When $\alpha = 1$, $A$ is an exact algorithm.

For worst-case $\alpha$-approximate algorithms, 
we can obtain a robustness result similar to Theorem \ref{theorem:average-active}.
The proof of the following theorem is given in Appendix \ref{apd-proof:theorem:worst-active}.

\begin{theorem}
\label{theorem:worst-active}
Assume $f_p$ is Lipschitz continuous with a Lipschitz constant $L$.
If $A$ is a worst-case $\alpha$-approximate algorithm, then
for any true prior $p_0$ and perturbed prior $p_1$, \\[2pt]
$\displaystyle {\hskip 2mm} f^{\worst}_{p_0}(A(p_1)) \geq \alpha \max_{\pi} f^{\worst}_{p_0}(\pi) - (\alpha + 1) L \| p_1 - p_0 \|$.
\end{theorem}

The worst-case utility is also computed with respect to the true prior $p_0$
(i.e., using $f^{\worst}_{p_0}$ instead of $f^{\worst}_{p_1}$).
Theorem \ref{theorem:worst-active} shows that when we use a perturbed prior,
the worst-case utility achieved by a worst-case $\alpha$-approximate algorithm degrades by at most a constant
factor of the $\ell_1$ distance between the perturbed prior and the true prior.

{\bf Application to Least Confidence:}
Theorem \ref{theorem:worst-active} implies the robustness of 
the well-known least confidence AL algorithm 
\cite{lewis1994sequential,culotta2005reducing} with perturbed priors.
This algorithm greedily selects 
the next example $x^*$ satisfying
${ x^* = \arg \min_x \{ \max_{y \in \labelset} p_{\mathcal{D}}[y;x] \} }$.
If $f_p$ is the version space reduction utility 
(considered previously for the maximum Gibbs error algorithm), 
we have the following corollary.
The proof of this corollary is given in Appendix \ref{apd-proof:corollary:least-conf}.

\begin{corollary}
\label{corollary:least-conf}
If $A$ is the least confidence algorithm, then for any true prior $p_0$ and
any perturbed prior $p_1$,
\begin{equation*}
f^{\worst}_{p_0}(A(p_1)) \ge \Big( 1 - \frac{1}{e} \Big) \max_{\pi} f^{\worst}_{p_0}(\pi) - \Big( 2 - \frac{1}{e} \Big) \| p_1 - p_0 \|.
\end{equation*}
\end{corollary}

{\bf Application to Generalized Maximum Gibbs Error:}
Theorem \ref{theorem:worst-active} also implies the 
robustness of the worst-case generalized Gibbs error algorithm
\cite{cuong2014near} with a bounded loss.
Intuitively, the algorithm greedily maximizes in the worst case the 
total generalized version space reduction,
which is defined as 
\begin{equation*}
t_p(S, h) \defeq {\hskip -2mm} \sum_{\substack{h',h'': h'(S) \ne h(S) \text{ or } \\ h''(S) \ne h(S)}} {\hskip -3mm} p[h'] \, L(h',h'') \, p[h''], 
\end{equation*}
where $L$ is a non-negative loss function between labelings that satisfies
$L(h, h')=L(h', h)$ and $L(h, h) = 0$ for all $h,h'$.
The worst-case generalized Gibbs error algorithm attempts to greedily maximize
$t^{\worst}_{p_0}(\pi) \defeq  \min_h t_{p_0}(x^\pi_h, h)$,
and it is a worst-case $(1-1/e)$-approximate algorithm for this objective \cite{cuong2014near}.

If we assume $L$ is upper bounded by a constant $m$, 
we have the following corollary about this algorithm.
The proof of this corollary is given in Appendix \ref{apd-proof:corollary:gengibbs}.
Note that the bounded loss assumption is reasonable since it
holds for various practical loss functions such as Hamming loss or $F_1$ loss,
which is $1 - F_1(h,h')$ where $F_1(h,h')$ is the $F_1$ score between $h$ and $h'$.

\begin{corollary}
\label{corollary:gengibbs}
If $A$ is the worst-case generalized Gibbs error algorithm and 
the loss function of interest is upper bounded by a constant $m \ge 0$, then
for any true prior $p_0$ and any perturbed prior $p_1$,
\begin{equation*}
t^{\worst}_{p_0}(A(p_1)) \ge \big( 1 - \frac{1}{e} \big) \max_{\pi} t^{\worst}_{p_0}(\pi) - m \big( 4 - \frac{2}{e} \big) \| p_1 - p_0 \|.
\end{equation*}
\end{corollary}

\subsection{Discussions}

We emphasize that our results are important as they enhance 
our understanding and confidence about existing AL algorithms.
Furthermore, if the utility we want to maximize is not Lipschitz continuous,
then even an exact AL algorithm for perturbed priors may not be robust,
both in the average and worst cases.
We prove this in Theorem \ref{theorem:counter-ex} below 
(see Appendix \ref{apd-proof:theorem:counter-ex} for proof).

\begin{theorem}
\label{theorem:counter-ex}
For both the average and worst cases, 
there is an AL problem with a non-Lipschitz utility such that:
for any $C, \alpha, \epsilon > 0$, there exist a perturbed prior $p_1$ 
satisfying ${ 0 < \| p_1 - p_0 \| < \epsilon }$ and an exact algorithm $A^*$ satisfying
\begin{equation*}
f^{c}_{p_0}(A^*(p_1)) < \alpha \max_{\pi} f^{c}_{p_0}(\pi) - C \| p_1 - p_0 \|,
\end{equation*}
where $f^{c}_{p_0} \in \{ f^{\avg}_{p_0}, f^{\worst}_{p_0}\}$ respectively.
\end{theorem}

This theorem and our results above suggest we should
use a Lipschitz utility for AL to maintain the robustness.

By taking $p_1 = p_0$, Corollaries \ref{corollary:maxGEC} and 
\ref{corollary:batch-maxGEC} can recover the approximation ratios 
for the maximum Gibbs error and batch maximum Gibbs error algorithms in 
Theorems 4 and 5 of \cite{cuong2013active} respectively.
Similarly, Corollaries \ref{corollary:least-conf} and \ref{corollary:gengibbs} 
can recover the ratios for the least confidence and generalized Gibbs error algorithms 
in Theorems 5 and 8 of \cite{cuong2014near} respectively.
Thus, our corollaries are generalizations of these previous theorems.

If $A$ is $\alpha$-approximate 
(in the average or worst case) with an optimal constant $\alpha$ 
under some computational complexity assumption \cite{golovin2011adaptive},
then it is also optimal in our theorems under the same assumption.
This can be proven easily by contradiction and setting $p_1 = p_0$.

If we are only interested in some particular prior $p_0$ and the perturbed priors
within a neighborhood of $p_0$, we can relax the Lipschitz assumption \eqref{eq:Lipschitz}
to the locally Lipschitz assumption at $p_0$: there exist $L$ and $\delta$
such that for all $S$, $h$, and $p$, if $\| p_0 - p \| < \delta$, then
$|f_{p_0}(S, h) - f_{p}(S, h)| \le L \|p_0 - p\|$.
Under this relaxed assumption, the theorems and corollaries above
still hold for any $p_1$ satisfying ${ \| p_0 - p_1 \| < \delta }$.

\section{Robustness: Minimum Cost Problem}

In this section, we investigate the robustness of AL algorithms 
for the \emph{minimum cost} problem in the average case:
find an adaptive policy minimizing the expected number of queries
to identify the true labeling $h_{\true}$ \cite{golovin2011adaptive}.
This problem assumes $h_{\true}$ is drawn from a prior $p_0$ on a small hypothesis space $\mathcal{H}$
(i.e., $\mathcal{H}$ does not need to contain all functions from $\itemset$ to $\labelset$).
After we make a query and observe a label, all the hypotheses inconsistent with the observed label are removed 
from the current hypothesis space (also called the version space).
We stop when there is only one hypothesis $h_{\true}$ left.

We do not consider this problem in the worst case because even the optimal
worst-case algorithm may not be robust.\footnote{The Lipschitz assumption is not satisfied in this setting.}
For instance, if the true prior gives probability 1 to one correct hypothesis 
but the perturbed prior gives positive probabilities to all the hypotheses,
then the cost of using the true prior is 0 while the cost of using the perturbed prior is $|\itemset|$.

For any policy $\pi$ and hypothesis $h$, let $c(\pi,h)$ be the cost of identifying $h$ when running $\pi$.
This is the length of the path corresponding to $h$ in the policy tree of $\pi$.
For any prior $p_0$ and policy $\pi$, the expected cost of $\pi$ with respect to the prior $p_0$ is defined as
$c^{\avg}_{p_0}(\pi) \defeq \mathbb{E}_{h \sim p_0} \left[ c(\pi, h) \right]$.

We will consider $\alpha(p)$-approximate algorithms that return a policy 
whose expected cost is within an $\alpha(p)$-factor of the optimal one.
Formally, for any prior $p$, an $\alpha(p)$-approximate (${\alpha(p) \ge 1}$) algorithm $A$ outputs a policy $A(p)$ satisfying
\begin{equation*} 
c^{\avg}_{p}(A(p)) \le \alpha(p) \min_{\pi} c^{\avg}_{p}(\pi).
\end{equation*}
Note that $\alpha(p)$ may depend on the prior $p$. 
When $\alpha(p) = 1$, $A$ is an exact algorithm.
We have the following robustness result for the minimum cost problem in the average case.
The proof of this theorem is given in Appendix \ref{apd-proof:theorem:min-cost}.

\begin{theorem}
\label{theorem:min-cost}
Assume $c(\pi,h)$ is upper bounded by a constant $K$ for all $\pi$, $h$.
If $A$ is an $\alpha(p)$-approximate algorithm, then for any true prior $p_0$
and any perturbed prior $p_1$, 
\begin{equation*}
c^{\avg}_{p_0}(A(p_1)) \le \alpha(p_1) \min_{\pi} c^{\avg}_{p_0}(\pi) + (\alpha(p_1) + 1) K \| p_1 - p_0 \|.
\end{equation*}
\end{theorem}

The assumption $c(\pi,h) \le K$ for all $\pi$, $h$ is reasonable 
since $c(\pi,h) \le |\itemset|$ for all $\pi$, $h$.
When $\mathcal{H}$ is small, $K$ can be much smaller than $|\itemset|$.

{\bf Application to Generalized Binary Search:}
Theorem \ref{theorem:min-cost} implies the robustness of the generalized binary search algorithm,
which is known to be $(\ln \frac{1}{\min_h p[h]}+1)$-approximate \cite{golovin2011adaptive}.
The result is stated in the corollary below.
By taking $p_1 = p_0$, this corollary can recover the previous result 
by \citeauthor{golovin2011adaptive} \shortcite{golovin2011adaptive} for the generalized binary search algorithm.

\begin{corollary}
\label{corollary:gbs}
Assume $c(\pi,h)$ is upper bounded by $K$ for all $\pi$, $h$.
If $A$ is the generalized binary search algorithm, then for any true prior $p_0$
and any perturbed prior $p_1$,
\begin{align*}
c^{\avg}_{p_0}(A(p_1)) \le& \left( \ln \frac{1}{\min_h p_1[h]}+1 \right) \min_{\pi} c^{\avg}_{p_0}(\pi) \\
 & + \left( \ln \frac{1}{\min_h p_1[h]}+2 \right) K \| p_1 - p_0 \|.
\end{align*}
\end{corollary}

Theorem \ref{theorem:min-cost} can also provide the robustness of algorithms 
for problems other than AL.
For instance, it can provide the robustness of the RAId algorithm for the 
adaptive informative path planning problem \cite{lim2015adaptive}.

\section{Mixture Prior}
\label{sec:mixture}

Let us consider methods that minimize a regularized loss. These methods are commonly used and known to be equivalent to finding the maximum a posteriori hypothesis with an appropriate prior. In practice, the best regularization constant is usually unknown, and a common technique (in passive learning) is to split the available data set into a training and a validation set, which is used to select the best regularization constant based on performance of the algorithm trained on the training set. 
As this method is effective in practice, we construct a Bayesian version and study its performance, particularly the robustness, when used with AL. We assume that we have a candidate set of prior distributions corresponding to different regularization constants, and the true hypothesis is randomly generated by first randomly selecting a distribution and then selecting a hypothesis using that distribution. This corresponds to assuming that the prior distribution is the mixture distribution. For simplicity, we consider the uniform mixture in this work.

First, we note that optimizing the expected cost of the mixture directly has a lower expected cost than trying to separately identify the appropriate component (corresponding to using a validation set in passive learning) and the best hypothesis given the component (corresponding to using the training set). Hence, we would expect the method to perform favorably in comparison to passive learning when the mixture prior is the true prior.

Results in earlier sections assure us that the method is near optimal when the mixture prior is incorrect but generates hypotheses with probabilities similar to the true prior. What if the true prior is far from the mixture distribution in the $\ell_1$ distance? In particular, we are interested in the case where the true distribution is one of the mixture components, rather than the mixture itself. Theorem \ref{theorem:mixture} below provides bounds on the performance in such cases (see Appendix \ref{apd-proof:theorem:mixture} for proof).
We note that the theorem holds for general priors that may vary in form (e.g., with different probability mass functions) and is not restricted to priors corresponding to regularization constants.

The bounds show that the performance of the mixture is competitive with that of the optimal algorithm, although the constant can be large if some hypotheses have small probabilities under the true distribution. 
We also provide an absolute bound (instead of competitive bound) which may be more informative in cases where there are hypotheses with small probabilities. The bound (the first bound in Theorem \ref{theorem:mixture}) shows that the expected cost is within a constant factor of the optimal expected cost of the mixture, which is the expected cost we would have to pay if our model was correct. The optimal expected cost of the mixture is in turn better than the expected cost of any two-stage identification procedure that identifies the component and the hypothesis given the component separately, assuming the expectation is taken with respect to the mixture.

\begin{theorem}
\label{theorem:mixture}
If $A$ is an $\alpha(p)$-approximate algorithm for the minimum cost problem, 
then for any true prior $p_0$ and any $k$ component uniform mixture prior $p_1 = \sum_{i=1}^k \frac{1}{k} p_{1,i}$
such that $p_0 \in \{ p_{1,i} \}_{i=1}^k$, we have
\[ c^{\avg}_{p_0}(A(p_1)) \le k\alpha(p_1) \min_{\pi} c^{\avg}_{p_1}(\pi), \text{ and} \]
\[ c^{\avg}_{p_0}(A(p_1)) \le \alpha(p_1) \Big( \frac{k-1}{\min_h p_0[h]} + 1 \Big) \min_{\pi} c^{\avg}_{p_0}(\pi). \]
As a result, if $A$ is generalized binary search, then \\[2pt]
${\hskip 3mm} c^{\avg}_{p_0}(A(p_1)) \le k \Big( \ln \frac{k}{\min_h p_0[h]} + 1 \Big) \, \displaystyle{ \min_{\pi} c^{\avg}_{p_1}(\pi) }$, and \\[3pt]
$c^{\avg}_{p_0}(A(p_1)) \le \big( \ln \frac{k}{\min_h p_0[h]} {\hskip -1mm} + {\hskip -1mm} 1 \big) \big(\frac{k-1}{\min_h p_0[h]} {\hskip -1mm}  + {\hskip -1mm}  1 \big) 
\displaystyle{ \min_{\pi} c^{\avg}_{p_0}(\pi) }.$
\end{theorem}

\begin{algorithm}[t]
\caption{Active learning for the mixture prior model}
\label{alg:mixture}
\begin{algorithmic}
   \STATE {\bfseries Input:} A set of $n$ priors $\{ p^1, p^2, \ldots, p^n \}$, 
      the initial normalized weights for the priors $\{ w^1, w^2, \ldots, w^n \}$,
      and the budget of $k$ queries. \\[4pt]
   \STATE $p^i_0 \gets p^i$; \quad $w^i_0 \gets w^i$; \quad for all $i = 1, 2, \ldots, n$; \\[4pt]
   \FOR{$t=1$ {\bfseries to} $k$}
      \STATE Choose an unlabeled example $x^*$ based on an AL criterion; \\
      \STATE $y^* \gets$ Query-label($x^*$); \\
      \STATE Update and normalize weights: \\
         ${\hskip 5mm} w^i_t \propto w^i_{t-1} \, p^i_{t-1}[y^*; x^*]$ \,\, for all $i = 1, 2, \ldots, n$; \\
      \STATE Update each posterior individually using Bayes' rule: \\
         ${\hskip 5mm} p^i_t[h] \propto p^i_{t-1}[h] \, \mathbb{P}[h(x^*) = y^* \mid h]$ \\
         {\hskip 2cm} for each $i = 1, 2, \ldots, n$ and $h \in \mathcal{H}$; \\
   \ENDFOR
   \STATE {\bfseries return} $\{ p^1_k, p^2_k, \ldots, p^n_k \}$
       and $\{ w^1_k, w^2_k, \ldots, w^n_k \}$;
\end{algorithmic}
\end{algorithm}

\begin{table*}[t!]
\centering{
\caption{AUCs of maximum Gibbs error algorithm with 
$1/\sigma^2 =$ 0.01, 0.1, 1, 10 and the mixture prior model on 20 Newsgroups data set (upper half) and UCI data set (lower half).
Double asterisks (**) indicate the best score, 
while an asterisk (*) indicates the second best score on a row (without the last column).
The last column shows the AUCs of passive learning with the mixture prior model for comparison.}
\label{tbl:auc-mix}
\vskip 3mm
\begin{tabular}{ccccccc}
\toprule
Data set & $0.01$ & $0.1$ & $1$ & $10$ & Mixture & Mixture (Passive) \\
\midrule
alt.atheism/comp.graphics & 87.60** & 87.25 & 84.94 & 81.46 & 87.33* & 83.92 \\
talk.politics.guns/talk.politics.mideast & 80.71** & 79.28 & 74.57 & 66.76 & 79.49* & 76.34 \\
comp.sys.mac.hardware/comp.windows.x & 78.75** & 78.21* & 75.07 & 70.54 & 78.21* & 75.02 \\
rec.motorcycles/rec.sport.baseball & 86.20** & 85.39 & 82.23 & 77.35 & 85.59* & 81.56 \\
sci.crypt/sci.electronics & 78.08** & 77.35 & 73.92 & 68.72 & 77.42* & 73.08 \\
sci.space/soc.religion.christian & 86.09** & 85.12 & 81.48 & 75.51 & 85.50* & 80.31 \\
soc.religion.christian/talk.politics.guns & 86.16** & 85.01 & 80.91 & 74.03 & 85.46* & 81.81 \\[2pt]
{\bf Average (20 Newsgroups)} & 83.37** & 82.52 & 79.02 & 73.48 & 82.71* & 78.86 \\
\midrule
\midrule
Adult \nocite{kohavi1996scaling} & 79.38 & 80.15 & 80.39** & 79.68 & 80.18* & 77.41 \\
Breast cancer \nocite{wolberg1990multisurface} & 88.28* & 88.37** & 86.95 & 83.82 & 88.14 & 89.07 \\
Diabetes \nocite{smith1988using} & 65.09* & 64.53 & 64.39 & 65.48** & 64.82 & 64.24 \\
Ionosphere \nocite{sigillito1989classification} & 82.80* & 82.76 & 81.48 & 77.88 & 82.95** & 81.91 \\
Liver disorders \nocite{forsyth1990pc} & 66.31** & 64.16 & 61.42 & 58.42 & 64.73* & 65.89 \\
Mushroom \nocite{schlimmer1987concept} & 90.73** & 89.56 & 84.14 & 82.94 & 90.33* & 73.38 \\
Sonar \nocite{gorman1988analysis} & 66.75** & 65.45* & 63.74 & 60.81 & 65.00 & 66.53 \\[2pt]
{\bf Average (UCI)} & 77.05** & 76.43 & 74.64 & 72.72 & 76.59* & 74.06 \\
\bottomrule
\end{tabular}
}
\end{table*}

The algorithm for greedy AL with the mixture model is shown in Algorithm~\ref{alg:mixture}. In the algorithm, the unlabeled example $x^*$ can be chosen using any AL criterion.
The criterion can be computed from the weights and posteriors 
obtained from the previous iteration.
For instance, if the maximum Gibbs error algorithm is used, then at iteration $t$, we have
${ x^* = \arg \max_x \mathbb{E}_{y \sim p[\cdot;x]} [ 1 - p[y;x] ] }$,
where ${ p[y;x] = \sum_{i=1}^n w^i_{t-1} \, p^i_{t-1}[y;x] }$.
After $x^*$ is chosen, we query its label $y^*$ and update 
the new weights and posteriors, which are always normalized so that 
$\sum_i w^i_t = 1$ for all $t$ and $\sum_h p^i_t[h] = 1$ for all $i$ and $t$.
The algorithm returns the final weights and posteriors that can be used
to make predictions on new examples.
More specifically, the predicted label of a new example $x$ is
$\arg \max_y \sum_{i=1}^n w^i_k \, p^i_k[y;x]$.

We note that Algorithm \ref{alg:mixture} does not require the hypotheses
to be deterministic. In fact, the algorithm can be used with probabilistic hypotheses
where $\mathbb{P}[h(x) = y \mid h] \in [0,1]$.
We also note that computing $p^i_t[y;x]$ for a posterior $p^i_t$ can be expensive. In this work, we approximate it using the MAP hypothesis.
In particular, we assume $p^i_t[y;x] \approx p^i_{\MAP}[y;x]$,
the probability that $x$ has label $y$ according to the MAP hypothesis of the 
posterior $p^i_t$.
This is used to approximate both the AL criterion
and the predicted label of a new example.

\section{Experiments}

We report experimental results with different priors and the mixture prior.
We use the logistic regression model with different $L_2$ regularizers,
which are well-known to impose a Gaussian prior with mean zero 
and variance $\sigma^2$ on the parameter space.
Thus, we can consider different priors by varying the variance $\sigma^2$
of the regularizer.
We consider two experiments with the maximum Gibbs error algorithm.
Since our data sets are all binary classification,
this algorithm is equivalent to the least confidence and the maximum entropy algorithms.
In our first experiment, we compare models that use different priors (equivalently, regularizers).
In the second experiment, we run the uniform mixture prior model
and compare it with models that use only one prior.
For AL, we randomly choose the first 10 examples as a seed set.
The scores are averaged over 100 runs of the experiments with different seed sets.

\subsection{Experiment with Different Priors}

\begin{figure}[t]
\centering
\includegraphics[height=3.1cm]{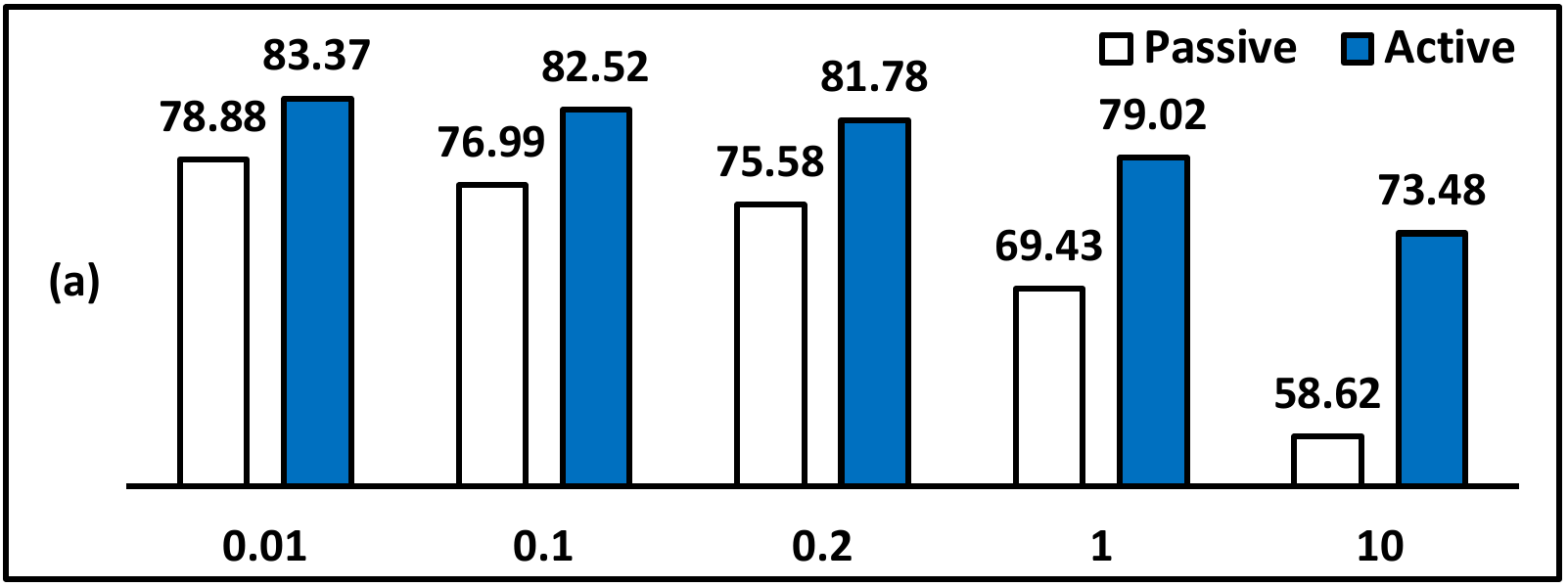}
\includegraphics[height=3.1cm]{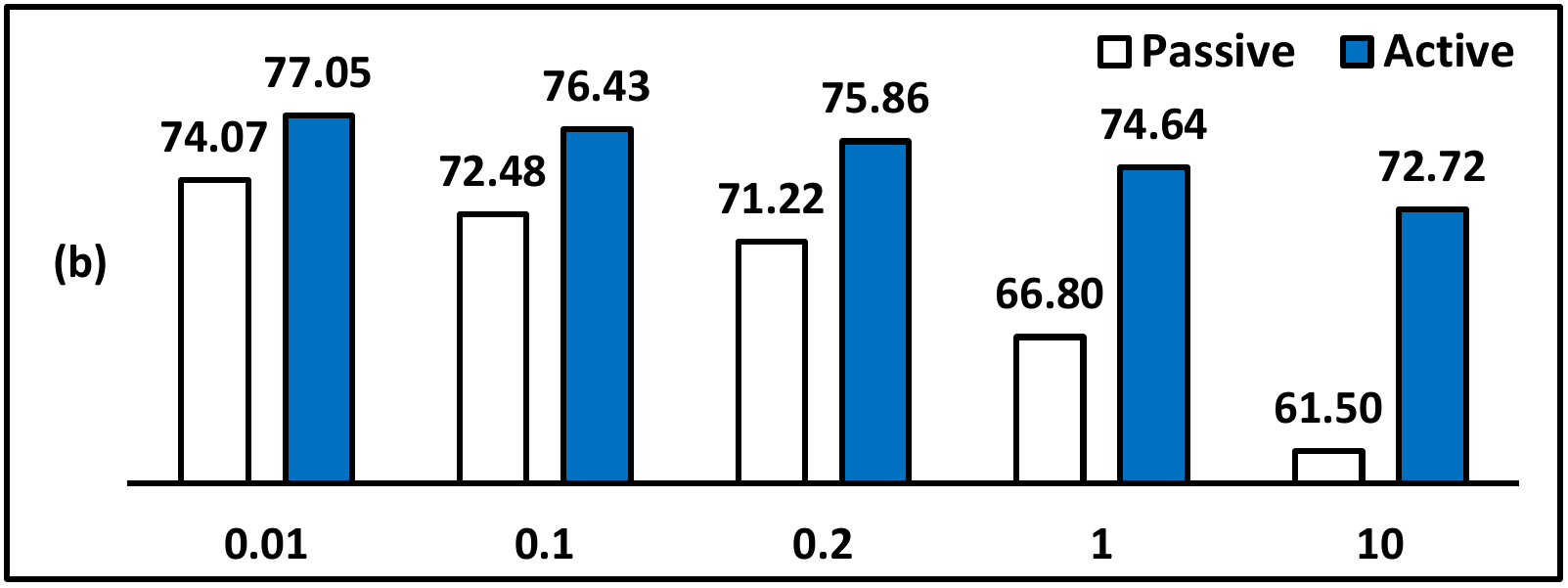}
\caption{Average AUCs for passive learning and maximum Gibbs error AL algorithms with $1/\sigma^2 =$ 0.01, 0.1, 0.2, 1, and 10 on the 20 Newsgroups (a) and UCI (b) data sets.}
\label{fig:regularizers}
\end{figure}

We run maximum Gibbs error with
$1/\sigma^2 = 0.01, \, 0.1, \, 0.2, \, 1, \, 10$
on tasks from the 20 Newsgroups
and UCI data sets \cite{joachims1996probabilistic,bache2013uci}
shown in the first column of Table \ref{tbl:auc-mix}.
Figure \ref{fig:regularizers} shows the average areas under the accuracy curves (AUC) 
on the first 150 selected examples for the different regularizers.
Figures \ref{fig:regularizers}a and \ref{fig:regularizers}b give the average AUCs (computed on a separate test set) for the 20 Newsgroups and UCI data sets respectively.
We also compare the scores for AL with passive learning.

From Figure \ref{fig:regularizers}, AL is better than passive learning for
all the regularizers. When the regularizers are close to each other 
(e.g., $1/\sigma^2 = 0.1$ and $0.2$), the corresponding scores tend to be close.
When they are farther apart (e.g., $1/\sigma^2 = 0.1$ and $10$),
the scores also tend to be far from each other.
In some sense, this confirms our results in previous sections.

\subsection{Experiment with Mixture Prior}

We investigate the performance of the mixture prior model
proposed in Algorithm \ref{alg:mixture}.
For AL, it is often infeasible to use a validation set to choose the regularizers beforehand because we do not initially have any labeled data.
So, using the mixture prior is a reasonable choice in this case.

We run the uniform mixture prior with regularizers ${ 1/\sigma^2 = 0.01, \, 0.1, \, 1, \, 10 }$
and compare it with models that use only one of these regularizers.
Table \ref{tbl:auc-mix} shows the AUCs of the first 150 selected examples 
for these models on the 20 Newsgroups and the UCI data sets.

From the results, the mixture prior model
achieves the second best AUCs for all tasks in the 20 Newsgroups data set.
For the UCI data set, the model achieves the best score on Ionosphere 
and the second best scores on three other tasks.
For the remaining three tasks, it achieves the third best scores.
On average, the mixture prior model achieves the second best scores for both data sets.
Thus, the model performs reasonably well given the fact that 
we do not know which regularizer is the best to use for the data.
We also note that if a bad regularizer is used (e.g., $1/\sigma^2 = 10$),
AL may be even worse than passive learning with mixture prior.

\section{Conclusion}
We proved new robustness bounds for AL with perturbed priors
that can be applied to various AL algorithms used in practice.
We showed that if the utility is not Lipschitz, an optimal algorithm on perturbed priors may not be robust.
Our results suggest that we should use a Lipschitz utility for AL if robustness is required.
We also proved novel robustness bounds for a uniform mixture prior
and showed experimentally that this prior is reasonable in practice.

\medskip\noindent{\bf Acknowledgments}.
We gratefully acknowledge the support of the Australian Research Council
through an Australian Laureate Fellowship (FL110100281) and
through the Australian Research Council Centre of Excellence for Mathematical
and Statistical Frontiers (ACEMS), and of QUT through a Vice Chancellor's
Research Fellowship.
We also gratefully acknowledge the support of Singapore MOE AcRF Tier Two grant R-265-000-443-112.

\section*{Appendix}
\renewcommand{\thesubsection}{\Alph{subsection}}

\subsection{Proof of Theorem \ref{theorem:average-active}}
\label{apd-proof:theorem:average-active}

Let $C = L + M$.
For any policy $\pi$, note that:
\begin{eqnarray*}
& & |f^{\avg}_{p_0}(\pi) - f^{\avg}_{p_1}(\pi)| \\
&=& | ( \sum_h p_0[h] f_{p_0}(x^\pi_h, h) - \sum_h p_0[h] f_{p_1}(x^\pi_h, h) ) \\
& & + (\sum_h p_0[h] f_{p_1}(x^\pi_h, h) -  \sum_h p_1[h] f_{p_1}(x^\pi_h, h)) | \\
&\le& C \| p_1 - p_0 \|,
\end{eqnarray*}
where the last inequality holds due to the Lipschitz continuity and
boundedness of the utility function $f_p$.
Thus, if $\pi_1 = \arg \max_{\pi} f^{\avg}_{p_1}(\pi)$ and $\pi_0 = \arg \max_{\pi} f^{\avg}_{p_0}(\pi)$, it follows that:
\begin{eqnarray*}
f^{\avg}_{p_1}(\pi_1) {\hskip -2mm} &\ge& {\hskip -2mm} f^{\avg}_{p_1}(\pi_0) \geq f^{\avg}_{p_0}(\pi_0) - C \| p_1 - p_0 \|, \text{ and} \\
f^{\avg}_{p_0}(\pi) {\hskip -2mm} &\ge& {\hskip -2mm} f^{\avg}_{p_1}(\pi) - C \| p_1 - p_0 \| \text{ for all } \pi.
\end{eqnarray*}
Hence,
\begin{eqnarray*}
f^{\avg}_{p_0}(A(p_1)) {\hskip -2mm} 
&\ge& {\hskip -2mm} f^{\avg}_{p_1}(A(p_1)) - C \| p_1 - p_0 \| \\
&\ge& {\hskip -2mm} \alpha f^{\avg}_{p_1}(\pi_1) - C \| p_1 - p_0 \| \\
&\ge& {\hskip -2mm} \alpha (f^{\avg}_{p_0}(\pi_0) - C \| p_1 - p_0 \|) - C \| p_1 - p_0 \| \\
&=& \alpha \max_{\pi} f^{\avg}_{p_0}(\pi) - C (\alpha + 1) \| p_1 - p_0 \|,
\end{eqnarray*}
where the first and third inequalities are from the above discussions
and the second inequality holds as $A$ is $\alpha$-approximate.

\subsection{Proof of Corollary \ref{corollary:maxGEC}}
\label{apd-proof:corollary:maxGEC}

\citeauthor{cuong2013active} \shortcite{cuong2013active} showed that 
the maximum Gibbs error algorithm provides a constant factor approximation 
to the optimal policy Gibbs error, 
which is equivalent to the expected version space reduction $f^{\avg}_{p}(\pi)$.
Formally, they showed that, for any prior $p$,
\[ f^{\avg}_{p}(A(p)) \ge \left( 1-\frac{1}{e} \right) \max_{\pi} f^{\avg}_{p}(\pi), \]
where $A$ is the maximum Gibbs error algorithm.
That is, the algorithm is average-case $(1-1/e)$-approximate.

Furthermore, the version space reduction utility is upper bounded by $M = 1$; 
and for any priors $p$, $p'$, we also have
\begin{eqnarray*}
& & |f_{p}(S,h) - f_{p'}(S,h)| \\
&=& |p'[h(S);S] - p[h(S);S]| \\
&=& |\sum_{h'} p'[h'] \, \mathbb{P}[h'(S)=h(S)|h'] \\
& & {\hskip 1cm} - \sum_{h'} p[h'] \, \mathbb{P}[h'(S)=h(S)|h']| \\
&\le& \| p - p' \|.
\end{eqnarray*}
Thus, the version space reduction utility is Lipschitz continuous with $L=1$ and is upper bounded by $M=1$.
Hence, Corollary \ref{corollary:maxGEC} follows 
from Theorem \ref{theorem:average-active}.

\subsection{Proof of Theorem \ref{theorem:worst-active}}
\label{apd-proof:theorem:worst-active}

Let ${ \pi_0 = \arg \max_{\pi} f^{\worst}_{p_0}(\pi) }$
and ${ \pi_1 = \arg \max_{\pi} f^{\worst}_{p_1}(\pi) }$.
We have 
$f^{\worst}_{p_1}(\pi_1) \ge f^{\worst}_{p_1}(\pi_0) = f_{p_1}(x^{\pi_0}_{h_0}, h_0)$,
where $h_0 = \arg \min_h f_{p_1}(x^{\pi_0}_h, h)$.
Using the Lipschitz continuity of $f_p$ and the definition
of $f^{\worst}_{p_0}$, we have 
\begin{eqnarray*}
f_{p_1}(x^{\pi_0}_{h_0}, h_0) &\ge& f_{p_0}(x^{\pi_0}_{h_0}, h_0) - L \| p_0 - p_1 \| \\
&\ge& \min_h f_{p_0}(x^{\pi_0}_h, h) - L \| p_0 - p_1 \| \\
&=& f^{\worst}_{p_0}(\pi_0) - L \| p_0 - p_1 \|.
\end{eqnarray*}
Thus, $f^{\worst}_{p_1}(\pi_1) \ge f^{\worst}_{p_0}(\pi_0) - L \| p_0 - p_1 \|$.

Let $\pi = A(p_1)$ and $h^* = \arg \min_h f_{p_0}(x^{\pi}_h, h)$.
We have
${ f^{\worst}_{p_0}(\pi) = \min_h f_{p_0}(x^{\pi}_h, h) = f_{p_0}(x^{\pi}_{h^*}, h^*) }$.
By the Lipschitz continuity of $f_p$, we have
\begin{eqnarray*}
f_{p_0}(x^{\pi}_{h^*}, h^*) &\ge& f_{p_1}(x^{\pi}_{h^*}, h^*) - L \| p_0 - p_1 \| \\
&\ge& \min_h f_{p_1}(x^{\pi}_h, h) - L \| p_0 - p_1 \| \\
&=& f^{\worst}_{p_1}(\pi) - L \| p_0 - p_1 \| \\
&\ge& \alpha \max_{\pi} f^{\worst}_{p_1}(\pi) - L \| p_0 - p_1 \| \\
&=& \alpha f^{\worst}_{p_1}(\pi_1) - L \| p_0 - p_1 \|,
\end{eqnarray*}
where the last inequality holds as $A$ is $\alpha$-approximate.
Using the inequality relating $f^{\worst}_{p_1}(\pi_1)$ and
$f^{\worst}_{p_0}(\pi_0)$ above, we now have
\begin{eqnarray*}
f^{\worst}_{p_0}(\pi) &\ge& \alpha ( f^{\worst}_{p_0}(\pi_0) - L \| p_0 - p_1 \| ) - L \| p_0 - p_1 \| \\
&=& \alpha \max_{\pi} f^{\worst}_{p_0}(\pi) - (\alpha + 1) L \| p_0 - p_1 \|.
\end{eqnarray*}

\subsection{Proof of Corollary \ref{corollary:least-conf}}
\label{apd-proof:corollary:least-conf}

\citeauthor{cuong2014near} \shortcite{cuong2014near} have shown that 
using the least confidence algorithm can achieve a constant factor approximation 
to the optimal worst-case version space reduction.
Formally, if $f_p(S, h)$ is the version space reduction utility 
(that was considered previously for the maximum Gibbs error algorithm),
then $f^{\worst}_{p}(\pi)$ is the worst-case version space reduction of $\pi$, 
and it was shown \cite{cuong2014near} that, for any prior $p$,
\[
f^{\worst}_{p}(A(p)) \ge \left( 1-\frac{1}{e} \right) \max_{\pi} f^{\worst}_{p}(\pi), 
\]
where $A$ is the least confidence algorithm.
That is, the least confidence algorithm is worst-case $(1-1/e)$-approximate.

Since the version space reduction utility is Lipschitz continuous with $L = 1$
as shown in the proof of Corollary \ref{corollary:maxGEC},
Corollary \ref{corollary:least-conf} follows from Theorem \ref{theorem:worst-active}.

\subsection{Proof of Corollary \ref{corollary:gengibbs}}
\label{apd-proof:corollary:gengibbs}

It was shown by \citeauthor{cuong2014near} \shortcite{cuong2014near} that, 
for any prior $p$,
\[ t^{\worst}_p(A(p)) \ge \left( 1-\frac{1}{e} \right) \max_{\pi} t^{\worst}_p(\pi), \]
where $A$ is the worst-case generalized Gibbs error algorithm.
That is, the worst-case generalized Gibbs error algorithm is worst-case $(1-1/e)$-approximate.

If we assume the loss function $L$ is upper bounded by a constant $m$,
then $t_p$ is Lipschitz continuous with $L = 2m$.
Indeed, for any $S,h,p$, and $p'$, we have
\begin{eqnarray*}
{\hskip -3mm}& & |t_p(S,h) - t_{p'}(S,h)| \\
{\hskip -3mm}&=& | \sum_{\substack{h'(S) \ne h(S) \text{ or } \\ h''(S) \ne h(S)}} {\hskip -6mm} L(h',h'') (p[h'] p[h''] - p'[h'] p'[h''])| \\
{\hskip -3mm}&\le& m \sum_{\substack{h'(S) \ne h(S) \text{ or } \\ h''(S) \ne h(S)}} |p[h'] p[h''] - p'[h'] p'[h'']| \\
{\hskip -3mm}&=& m \sum_{\substack{h'(S) \ne h(S) \text{ or } \\ h''(S) \ne h(S)}} | (p[h'] - p'[h'])p[h''] \\
{\hskip -3mm}&& {\hskip 2.5cm} + p'[h'](p[h''] - p'[h''])| \\
{\hskip -3mm}&\le& m \sum_{h',h''} (|p[h']-p'[h']|p[h''] + p'[h']|p[h'']-p'[h'']|) \\
{\hskip -3mm}&=& 2 m \| p - p' \|.
\end{eqnarray*}
Thus, Corollary \ref{corollary:gengibbs} 
follows from Theorem \ref{theorem:worst-active}.

\subsection{Proof of Theorem \ref{theorem:counter-ex}}
\label{apd-proof:theorem:counter-ex}
For both the average and worst cases, consider the AL problem with budget $k=1$
and the utility
\[ f_p(S,h) = | \{ h' : p[h'] > \mu \text{ and } h'(S) \ne h(S) \} |, \]
for some very small $\mu > 0$ in the worst case and $\mu = 0$ in the average case.

This utility returns the number of hypotheses 
that have a significant probability (greater than $\mu$) 
and are not consistent with $h$ on $S$.
When $\mu=0$, it is the number of hypotheses pruned from the version space.
So, this is a reasonable utility to maximize for AL.
It is easy to see that this utility is non-Lipschitz.

Consider the case where there are two examples $x_0$, $x_1$ 
and 4 hypotheses $h_1, \ldots, h_4$ with binary labels given according to 
the following table.

\begin{table}[h]
\centering{
\begin{tabular}{ccc}
\toprule
Hypothesis & $x_0$ & $x_1$ \\
\midrule
$h_1$ & 0 & 0 \\
$h_2$ & 1 & 0 \\
$h_3$ & 0 & 1 \\
$h_4$ & 1 & 1 \\
\bottomrule
\end{tabular}
}
\end{table}

Consider the true prior $p_0$ where 
$p_0[h_1] = p_0[h_2] = \frac{1}{2} - \mu$ 
and $p_0[h_3] = p_0[h_4] = \mu$,
and a perturbed prior $p_1$ where
$p_1[h_1] = p_1[h_2] = \frac{1}{2} - \mu - \delta$ 
and $p_1[h_3] = p_1[h_4] = \mu + \delta$,
for some small $\delta > 0$.

With budget $k=1$, there are two possible policies: 
the policy $\pi_0$ which chooses $x_0$
and the policy $\pi_1$ which chooses $x_1$.
Let $A^*(p_1) = \pi_1$.
Note that
$f^{\avg}_{p_1}(\pi_1) = 2 (\frac{1}{2} - \mu - \delta) + 2 (\frac{1}{2} - \mu - \delta) + 2 (\mu + \delta) + 2 (\mu + \delta) = 2$,
and
$f^{\avg}_{p_1}(\pi_0) = 2 (\frac{1}{2} - \mu - \delta) + 2 (\frac{1}{2} - \mu - \delta) + 2 (\mu + \delta) + 2 (\mu + \delta) = 2$.
Thus, $\pi_1$ is an average-case optimal policy for $p_1$ and 
$A^*$ is an exact algorithm for $p_1$ in the average case.

Similarly, $f^{\worst}_{p_1}(\pi_1) = 2 = f^{\worst}_{p_1}(\pi_0)$.
Thus, $\pi_1$ is a worst-case optimal policy for $p_1$ and 
$A^*$ is an exact algorithm for $p_1$ in the worst case.
Hence, $A^*$ is an exact algorithm for $p_1$ in both average and worst cases.

Considering $p_0$, we have 
$f^{\avg}_{p_0}(\pi_1) = 0 (\frac{1}{2} - \mu) + 0 (\frac{1}{2} - \mu) + 2 \mu + 2 \mu = 0$ 
since $\mu = 0$ in the average case.
On the other hand,
$f^{\avg}_{p_0}(\pi_0) = 1 (\frac{1}{2} - \mu) + 1 (\frac{1}{2} - \mu) + 1 \mu + 1 \mu = 1$.
Similarly, in the worst case, we also have $f^{\worst}_{p_0}(\pi_1) = 0$ and $f^{\worst}_{p_0}(\pi_0) = 1$.
Thus, $\pi_0$ is the optimal policy for $p_0$ in both average and worst cases.
Now given any $C, \alpha, \epsilon > 0$, we can choose a small enough $\delta$
such that $\| p_1 - p_0 \| < \epsilon$ and $\alpha - C \| p_1 - p_0 \| > 0$.
Hence, Theorem \ref{theorem:counter-ex} holds.

\subsection{Proof of Theorem \ref{theorem:min-cost}}
\label{apd-proof:theorem:min-cost}

For any policy $\pi$, note that
\begin{eqnarray*}
|c^{\avg}_{p_0}(\pi) - c^{\avg}_{p_1}(\pi)| {\hskip -2mm} 
&=& {\hskip -2mm} | \sum_h p_0[h] c(\pi, h) - \sum_h p_1[h] c(\pi, h) | \\
&=& {\hskip -2mm} | \sum_h (p_0[h] - p_1[h]) c(\pi, h) | \\
&\le& {\hskip -2mm} K \| p_0 - p_1 \|,
\end{eqnarray*}
where the last inequality holds as $c(\pi, h)$ is upper bounded by $K$.
Thus,
\[ c^{\avg}_{p_0}(\pi) \le c^{\avg}_{p_1}(\pi) + K \| p_0 - p_1 \|, \text{ for all } \pi, p_0, p_1. \]

Let $\pi_0 = \arg \min_{\pi} c^{\avg}_{p_0}(\pi)$.
We have
\begin{eqnarray*}
c^{\avg}_{p_0}(A(p_1)) {\hskip -2mm}
&\le& {\hskip -2mm} c^{\avg}_{p_1}(A(p_1)) + K \| p_0 - p_1 \| \\
&\le& {\hskip -2mm} \alpha(p_1) \min_{\pi} c^{\avg}_{p_1}(\pi) + K \| p_0 - p_1 \| \\
&\le& {\hskip -2mm} \alpha(p_1) c^{\avg}_{p_1}(\pi_0) + K \| p_0 - p_1 \| \\
&\le& {\hskip -2mm} \alpha(p_1) (c^{\avg}_{p_0}(\pi_0) + K \| p_0 - p_1 \|) + K \| p_0 - p_1 \| \\
&=& {\hskip -2mm} \alpha(p_1) c^{\avg}_{p_0}(\pi_0) + (\alpha(p_1) + 1) K \| p_0 - p_1 \| \\
&=& {\hskip -2mm} \alpha(p_1) \min_{\pi} c^{\avg}_{p_0}(\pi) + (\alpha(p_1) + 1) K \| p_0 - p_1 \|,
\end{eqnarray*}
where the first and fourth inequalities are from the discussion above, 
and the second inequality is from the fact that $A$ is $\alpha(p)$-approximate.

\subsection{Proof of Theorem \ref{theorem:mixture}}
\label{apd-proof:theorem:mixture}

If $k=1$, then $p_1 = p_0$ and Theorem \ref{theorem:mixture} trivially holds.
Consider $k \ge 2$.
For any $h$, since 
\[ \frac{p_0[h]}{p_1[h]} = \frac{p_0[h]}{\sum_{i=1}^k \frac{1}{k} p_{1,i}[h]} \le \frac{p_0[h]}{\frac{1}{k} p_0[h]} = k, \]
we have
$k - 1 \ge 1 - \frac{p_0[h]}{p_1[h]} \ge 1 - k$.
Thus, $| 1 - \frac{p_0[h]}{p_1[h]} | \le k - 1$. \\

Hence, for any policy $\pi$,
\begin{eqnarray*}
|c^{\avg}_{p_1}(\pi) - c^{\avg}_{p_0}(\pi)|
&=& | \sum_h p_1[h] ( 1 - \frac{p_0[h]}{p_1[h]}) c(\pi, h) | \\
&\le& (k-1) \sum_h p_1[h] c(\pi, h) \\
&=& (k-1) c^{\avg}_{p_1}(\pi).
\end{eqnarray*}
Therefore, $c^{\avg}_{p_0}(\pi) \le k c^{\avg}_{p_1}(\pi)$. \\

On the other hand, for any $h$, we have
\begin{eqnarray*}
\frac{p_1[h]}{p_0[h]} {\hskip -2mm} 
&=& {\hskip -2mm} \frac{\sum_{i=1}^k \frac{1}{k} p_{1,i}[h]}{p_0[h]} = \frac{\frac{1}{k} p_0[h] + \sum_{i:p_{1,i} \ne p_0} \frac{1}{k} p_{1,i}[h] }{p_0[h]} \\
&\le& \frac{1}{k} + \frac{\frac{k-1}{k}}{\min_h p_0[h]} = \frac{1}{k} + \frac{k-1}{k \min_h p_0[h]}.
\end{eqnarray*}
Thus, $1 - \frac{1}{k} \ge 1 - \frac{p_1[h]}{p_0[h]} \ge 1 - \frac{1}{k} - \frac{k-1}{k \min_h p_0[h]}$.
When $\mathcal{H}$ contains at least 2 hypothesis, $\min_h p_0[h] \le 1/2$, and 
$\frac{1}{k} + \frac{k-1}{k \min_h p_0[h]} - 1 \ge 1 - \frac{1}{k}$
(the case when $\mathcal{H}$ is a singleton is equivalent to $k=1$).
Hence, 
\[| 1 - \frac{p_1[h]}{p_0[h]} | \le \frac{1}{k} + \frac{k-1}{k \min_h p_0[h]} - 1. \]

We have
\begin{eqnarray*}
|c^{\avg}_{p_0}(\pi) - c^{\avg}_{p_1}(\pi)| {\hskip -2mm} 
&=& {\hskip -2mm} | \sum_h p_0[h] ( 1 - \frac{p_1[h]}{p_0[h]}) c(\pi, h) | \\
&\le& {\hskip -2mm} (\frac{1}{k} + \frac{k-1}{k \min_h p_0[h]} - 1) c^{\avg}_{p_0}(\pi).
\end{eqnarray*}
Therefore, $c^{\avg}_{p_1}(\pi) \le (\frac{1}{k} + \frac{k-1}{k \min_h p_0[h]}) c^{\avg}_{p_0}(\pi)$.

Now let $\pi_0 = \arg \min_{\pi} c^{\avg}_{p_0}(\pi)$.
We have
\begin{eqnarray*}
c^{\avg}_{p_0}(A(p_1)) {\hskip -2mm}
&\le& {\hskip -2mm} k c^{\avg}_{p_1}(A(p_1)) \\
&\le& {\hskip -2mm} k \alpha(p_1) \min_{\pi} c^{\avg}_{p_1}(\pi) \text{\qquad \quad (first part)} \\
&\le& {\hskip -2mm} k \alpha(p_1) c^{\avg}_{p_1}(\pi_0) \\
&\le& {\hskip -2mm} k \alpha(p_1) (\frac{1}{k} + \frac{k-1}{k \min_h p_0[h]}) c^{\avg}_{p_0}(\pi_0) \\
&=& {\hskip -2mm} \alpha(p_1) (1 + \frac{k-1}{\min_h p_0[h]}) \min_{\pi} c^{\avg}_{p_0}(\pi),
\end{eqnarray*}
where the first and fourth inequalities are from the discussions above,
and the second inequality is from the fact that $A$ is $\alpha(p)$-approximate.

If $A$ is the generalized binary search algorithm, then 
$\alpha(p_1) = \ln \frac{1}{\min_h p_1[h]}+1$.
Note that $\min_h p_1[h] = \min_h \sum_{i=1}^k \frac{1}{k} p_{1,i}[h] \ge \min_h \frac{1}{k} p_0[h]$.
Thus, 
${ \alpha(p_1) \le \ln \frac{k}{\min_h p_0[h]}+1 }$.
Therefore, \\
${ c^{\avg}_{p_0}(A(p_1)) \le (\ln \frac{k}{\min_h p_0[h]} {\hskip -1mm} + {\hskip -1mm} 1) (\frac{k-1}{\min_h p_0[h]} {\hskip -1mm} + {\hskip -1mm} 1) \displaystyle{ \min_{\pi} c^{\avg}_{p_0}(\pi) } }$.

\bibliography{robust}
\bibliographystyle{aaai}

\end{document}